\documentclass[11pt,a4paper]{article}
\usepackage[hyperref]{emnlp2020}
\usepackage{times}
\usepackage{latexsym}
\usepackage{tabu,multirow}
\usepackage{graphicx}
\usepackage{array}
\usepackage{todonotes}
\usepackage[export]{adjustbox}

\usepackage[normalem]{ulem}

\usepackage{microtype}

\aclfinalcopy 

\title{Generating similes \sout{effortlessly} \textit{like a Pro}: \\A Style Transfer Approach for Simile Generation}

\author{Tuhin Chakrabarty\textsuperscript{1,2}\thanks{~~The research was conducted when the author was at USC/ISI.}, 
  Smaranda Muresan\textsuperscript{2,4}
  \textbf{and} \textbf{Nanyun Peng}\textsuperscript{1,3}\\ 
  \textsuperscript{1}Information Sciences Institute, University of Southern California \\
  \textsuperscript{2}Department of Computer Science, Columbia University \\
   \textsuperscript{3}Computer Science Department, University of California, Los Angeles \\
  \textsuperscript{4}Data Science Institute, Columbia University\\\AND
  {\tt \{tuhin.chakr, smara\}@cs.columbia.edu}\\
  {\tt violetpeng@cs.ucla.edu}
  }

\date{}

\begin{document}
\maketitle
\begin{abstract}
Literary tropes, from poetry to stories, are at the crux of human imagination and communication. Figurative language such as a simile go beyond plain expressions to give readers new insights and inspirations. In this paper, we tackle the problem of simile generation. Generating a simile requires proper understanding for effective mapping of properties between two concepts. To this end, we first propose a method to automatically construct a parallel corpus by transforming a large number of similes collected from Reddit to their literal counterpart using structured common sense knowledge. We then propose to fine-tune a pretrained sequence to sequence model, BART~\cite{lewis2019bart}, on the literal-simile pairs to gain generalizability, so that we can generate novel similes given a literal sentence. Experiments show that our approach generates $88\%$ novel similes that do not share properties with the training data. Human evaluation on an independent set of literal statements shows that our model generates similes better than two literary experts \textit{37\%}\footnote{We average 32.6\% and 41.3\% for 2 humans.} of the times, and three baseline systems including a recent metaphor generation model \textit{71\%}\footnote{We average 82\% ,63\% and 68\% for three baselines.} of the times when compared pairwise.\footnote{The simile in the title is generated by our best model. Input: Generating similes effortlessly, output: Generating similes \textit{like a Pro}.} We also show how replacing literal sentences with similes from our best model in machine generated stories improves evocativeness and leads to better acceptance by human judges.
\end{abstract}

\section{Introduction}

\begin{figure}[ht]
\centering
\includegraphics[width=\columnwidth]{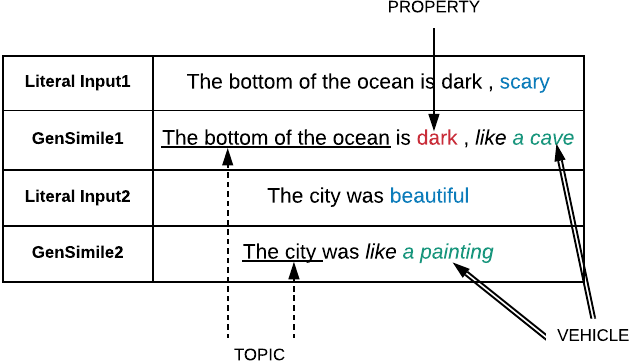}
\caption{\label{figure:simexample} Examples of two generated similes GenSimile1 and GenSimile2 from their literal inputs.} 
\vspace{-1em}
\end{figure}

Comparisons are inherent linguistic devices that express the likeness of two entities, concepts or ideas. 
When used in a figurative sense, these comparisons are called similes. They are a figure of speech that compare two different kind of things, usually with the intent to make the description more emphatic or vivid, being often used in literature and poetry to spark the reader's imagination ~\cite{definition}. Take the following two examples: ``The city was \textit{like a painting}", and ``If it falls into the wrong hands it would be as catastrophic \textit{as a nuclear bomb}." In the first example, the comparison draws on the implicit ``beauty" property being shared by the two very different entities, \textit{city} and \textit{painting}, while in the second the ``catastrophic" property is shared by \textit{falling into the wrong hands} and \textit{nuclear bomb}.

While most computational work has focused on simile detection \cite{simile1,simile2,simile3,simile4,simile5,simile6}, research on simile generation is under-explored. Generating similes could impact many downstream applications such as creative writing assistance, and literary or poetic content creation. To tackle the generation problem, we take advantage of the relatively simple structure of similes that consists of five elements \cite{hanks2013lexical,simile1}: the {\tt TOPIC} (usually a noun phrase that acts as logical subject), the {\tt VEHICLE} (the logical object of the comparison, usually a noun phrase), the {\tt PROPERTY} (what the two things being compared have in common, usually an adjective), the {\tt EVENT} (eventuality or state, usually a verb), and the {\tt COMPARATOR} (the trigger word or phrase that marks the presence of a comparison, usually the preposition ``like" or ``as...as"). All elements of a simile are explicit, with the exception of {\tt PROPERTY}, which can be both implicit and explicit. If we take the first example above, its structure is: ``[The city/{\tt TOPIC}] [was/{\tt EVENT}] [like/{\tt COMPARATOR}] [a painting/{\tt VEHICLE}]" ({\tt PROPERTY} is implicit). Unlike metaphors, the semantic context of similes tends to be very shallow, transferring a single \textit{property} \cite{hanks2013lexical}. Moreover, the  explicit syntactic structure of similes allows, in exchange, for more lexical creativity~\cite{simile1}. 

We focus on the task of generating a simile starting from a literal utterance that contains the {\tt TOPIC}, {\tt EVENT} and {\tt PROPERTY}. We frame this task as a style-transfer problem \cite{shen2017style,fu2017style,li2018delete,sudhakar2019transforming}, where the author's intent is to make the description of the {\tt TOPIC} more emphatic by introducing a comparison with the {\tt VEHICLE} via a shared {\tt PROPERTY} (See Figure~\ref{figure:simexample} for example of literal descriptive sentences and the generated similes). We call our approach \textbf{SCOPE} (\textbf{S}tyle transfer through \textbf{CO}mmonsense \textbf{P}rop\textbf{E}rty). There are two main challenges we need to address: 1) the lack of training data that consists of pairs of literal utterances and their equivalent simile in order to train a supervised model; 
2) ensuring that the generated simile makes a meaningful comparison between the {\tt TOPIC} and the {\tt VEHICLE} via the shared {\tt PROPERTY} explicitly or implicitly expressed (e.g., Figure  \ref{figure:simexample} GenSimile1 and GenSimile2, respectively). 
To the best of our knowledge, this is the first work in attempting to generate similes. By framing the task as a style-transfer problem we make three contributions:
\footnote{
Code \& Data at \url{https://github.com/tuhinjubcse/SimileGeneration-EMNLP2020}}



\textbf{Automatic creation of a parallel corpus of \textit{[literal sentence, simile]} pairs}. Our constructed corpus contains 87,843 such pairs. As a first step, we use distant supervision to automatically collect a set of \emph{self-labeled similes} using the phrase \textit{like a}. We then convert these similes to their literal versions by removing the {\tt COMPARATOR} and replacing the {\tt VEHICLE} with the associated {\tt PROPERTY} by leveraging the structured common sense knowledge achieved from COMET \cite{comet}, a language model fine-tuned on ConceptNet \cite{conceptnet}. For example, for the simile ``Love is like a unicorn" our method will generate ``Love is rare" (Section \ref{section:data1}).



\textbf{Transfer learning from a pre-trained model for generating high quality similes.} 
Our system \textbf{SCOPE}, 
\emph{fine-tunes} BART \cite{lewis2019bart} --- a state of the art pre-trained denoising autoencoder built with a sequence to sequence model, on our \emph{automatically collected parallel corpus} of \textit{[literal sentence, simile]} pairs (Section \ref{section:model}) to generate similes. Human evaluations show that this approach generates similes that are better 37\% of the time on average compared to 2 literary experts, 82\% and 63\%  of times compared to two well crafted baselines, and 68\% of the times compared to a state of the art system for metaphor generation \cite{metagen2} (Section \ref{section:results}). 

\textbf{A task-based evaluation.} We show the effectiveness of the generated similes as a tool for enhancing creativity and evocativeness in machine generated stories. Evaluation via Amazon Mechanical Turk shows that stories containing similes generated by \textbf{SCOPE} is preferred by Turkers 42\% of the times compared to stories without similes, which is preferred 25\% of the times (Section \ref{section:story}).

\section{SCOPE: {S}tyle Transfer through \textbf{CO}mmonsense \textbf{P}rop\textbf{E}rty}

\begin{figure*}[ht]
\centering
\includegraphics[width=\textwidth]{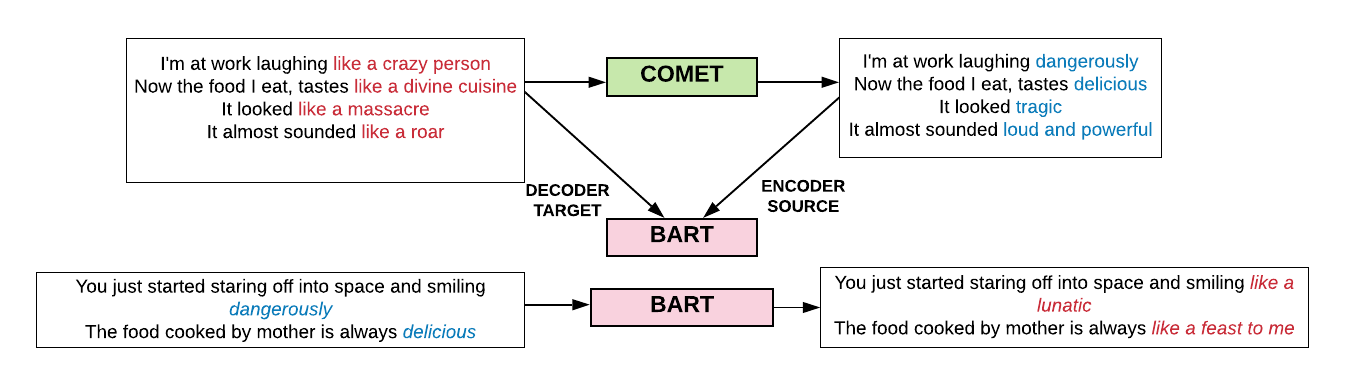}
\vspace{-1.5em}
\caption{\label{figure:sim} A schematic illustration of our system, where the top block shows our \textbf{training} process where we use COMET to transform similes to literal sentences and use them to fine-tune BART. The block below shows the \textbf{inference} step where we use fine-tuned BART to generate novel similes conditioned on a literal sentence.}
\vspace{-.5em}
\end{figure*}

Our style transfer approach for simile generation from literal descriptive sentences has two steps: 1) first convert self-labeled similes into literal sentences using structured common sense knowledge (Section \ref{section:data1}); and 2) given the \textit{[literal sentence, simile]} pairs, fine-tune a seq2seq model on these pairs to generate a simile given a literal sentence (Section \ref{section:model}). This two-step approach is shown in the upper half of Figure \ref{figure:sim}.

\subsection{Automatic Parallel Corpus Creation} \label{section:data1}
One of the requirements to train a supervised generative model for text style transfer is the presence of a large-scale parallel corpus. We use distant supervision to collect self-labeled similes using the phrase \textit{like a} \footnote{While there can be noisy sentences where the TOPIC is a PNP and typically short $<=6$ tokens such as \textit{I feel like a .., I would like a .., I don't like a..}, they are very less in number(1.1 \%) so we do not remove them. More details in Appendix A.2} from Reddit (e.g., the rows labeled as Simile in Table \ref{table:example2}). For fine-tuning, the similes form the ``target" side of our parallel data. For the ``source" side of our parallel data, we use commonsense knowledge to transform the similes to their literal version (e.g., the rows labeled as Best Literal in Table \ref{table:example2}). One of the possible ways to collect similes would be to train a supervised model using existing data and methods for simile detection but most data sets are very small in size (in order of a few hundred). The only large-scale dataset is that of \cite{simile1} however their data is from a rather restricted domain of product reviews on Amazon which might often lack variety, diversity and creativity needed for this task.

\paragraph{Simile Dataset Collection.}
We hypothesize that similes are used frequently in creative writing or humorous content on social media \cite{veale2013humorous}. Hence, we obtain training data by scraping the subreddits WRITINGPROMPTS \footnote{\url{https://www.reddit.com/r/WritingPrompts/}} and FUNNY \footnote{\url{https://www.reddit.com/r/funny/}} from social media site Reddit for comments containing the
phrase \textit{like a}. Similes can be both Open and Closed. For example the Closed Simile, \textit{``The boy was as strong as an ox"} gives \textbf{strong} as the PROPERTY shared by the \textit{boy} and \textit{ox}. But most similes do not give an explicit PROPERTY such as the Open Simile (e.g., \textit{``The boy was like an ox")} leaving the reader to infer that the boy is strong/large/fast \cite{simile4}. Due to their implicit nature, generating open similes is often more challenging and hence we resort to only using \textit{like} as a comparator instead of \textit{as...as}. We use the
API provided by  pushshift.io \footnote{\url{https://pushshift.io/}} to mine comments. Through this process we collect 87,843 self-labeled human written similes, from which we use 82,697 samples for training and 5,146 for validation. 



\paragraph{Simile to Literal Transformation via Commonsense Property.}
From a theoretical perspective, similes are created by making a comparison between the {\tt TOPIC} and the {\tt VEHICLE} through a shared {\tt PROPERTY}. While this property is naturally known to humans through common sense and connotative knowledge, computers still struggle to perform well on such tasks when the {\tt PROPERTY} is not expressed. Hence we use structured common sense knowledge to derive properties to transform similes to their literal versions.

\begin{table}[t]
\centering
\small
\begin{tabular}{|@{ }l@{ }|@{ }p{5.5cm}@{ }|}
\hline
Simile       & Love is like a \textit{unicorn.}                                                 \\ \hline
Has property     & very rare,  rare,  beautiful, beautiful and smart, color                  \\ \hline
Best Literal & Love is \textit{\color{blue}rare.}                                                           \\ \hline\hline
Simile       & It was cool and quiet, and I stormed through like a \textit{charging bull.}      \\ \hline
Has property     & big and strong, dangerous, big, fast, large                             \\ \hline
Best Literal & It was cool and quiet, and I stormed through \textit{\color{blue}fast.}                      \\ \hline\hline
Simile       & Sir Francis's voice was calm and quiet, like a \textit{breeze through a forest.}  \\ \hline
Has property     & very relax, soothe, cool, beautiful, relax                              \\ \hline
Best Literal & Sir Francis's voice was calm and quiet, \textit{\color{blue}very relaxed.}                   \\ \hline
\end{tabular}
\caption{Examples of self-labeled similes collected from Reddit. For each example, we show the top five commonsense properties associated with the \textit{vehicle} obtained from COMET, and the best literal sentence constructed from these properties. The blue italic texts in the literal sentences represent the \textit{property} inferred from the \textit{vehicle} in the simile (denoted in black italic). }
\vspace{-1em}
\label{table:example2}
\end{table}

To generate the common sense {\tt PROPERTY} that is implied by the {\tt VEHICLE} in the simile, we take advantage of the simple syntactic structure of a simile. We extract the {\tt VEHICLE} by extracting the phrase after \textit{like a} and feed it as input to COMET \cite{comet}. COMET is an adaptation framework for constructing commonsense knowledge based on pre-trained language models. Our work only leverages the \textbf{HasProperty} relation from COMET \footnote{\url{https://mosaickg.apps.allenai.org/comet\_conceptnet}}.


For a given simile \textit{`Love is like a unicorn.'}, the {\tt TOPIC} \textit{Love} is compared to the {\tt VEHICLE} \textit{unicorn}. As shown in Table \ref{table:example2}, COMET tells us the top 5 properties associated with the {\tt VEHICLE} are \textit{very rare, rare, beautiful, beautiful and smart, color}.

COMET gives us the properties sorted by probability in isolation by just relying on the {\tt VEHICLE}. While in most situations all of the properties are apt, we need to make the literal sentence as meaningful as possible. To do this, we append the common sense property to the portion of the simile before \textit{`like a'}. This typically consists of the {\tt TOPIC}, the {\tt EVENT}, and a {\tt PROPERTY} if stated explicitly. We take the top 5 properties from COMET to form 5 possible literal versions for a particular simile. To rank these literal versions and select the best one, we rely on perplexity scores obtained from a pre-trained language model GPT \cite{gpt}. Table \ref{table:example2} shows human written similes collected from Reddit, the top 5 common sense properties associated with the {\tt VEHICLE}, and the literal version created by taking the best {\tt PROPERTY}. To correct any grammatical error introduced by this manipulation, we rely on a grammatical error correction model \cite{zhao2019improving}.


\paragraph{Test Data Collection.} \label{section:evaldata}
Our task is to generate a simile given a literal input. The automatically-generated parallel data  might contain stylistic biases. To truly measure the effectiveness of our approach, we need to evaluate on a dataset independent of our training and validation data. Towards this end, we again scrape WRITINGPROMPTS subreddits for sentences which are this time \emph{literal} in nature (without any comparators \textit{like, as}). Since literal utterances contains the description of {\tt TOPIC} via a {\tt PROPERTY} and usually the {\tt PROPERTY} is an adjective or adverb, 
we restrict the last word of our literal sentences to adverbs or adjectives. We crawl 500 such sentences and randomly sample 150 literal utterance. We used two literary experts, a student in creative writing, and a student in comparative literature who is the author of a novel, to write corresponding similes for each of these 150 inputs for evaluation and comparison.

\subsection{Seq2Seq Model for Simile Generation} \label{section:model}
Our goal of generating similes can be broken down into two primary tasks: 1) identifying the words in the literal sentence that should be removed or replaced and 2) generating the appropriate substitutions while being pertinent to the context. Sequence to sequence (seq2seq) neural network models
\cite{sutskever2014sequence} have demonstrated great
success in many text generation tasks, such as machine translation, dialog system and image caption, with the requirement of a considerable amount of parallel data. Hence we use seq2seq models for simile generation.

\begin{figure}[t]
\centering
\includegraphics[scale=0.75]{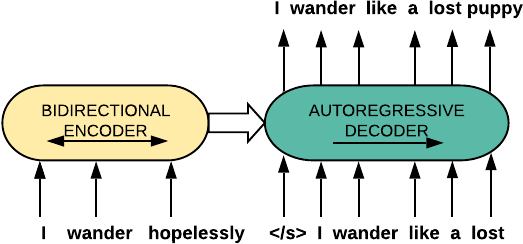}
\caption{\label{figure:bart} The backbone of SCOPE:  
fine-tuning BART on literal to simile pairs.} 
\vspace{-1em}
\end{figure}

BART \cite{lewis2019bart} is a pre-trained model combining bidirectional and auto-regressive transformers. It is implemented as a sequence-to-sequence model with a bidirectional encoder over corrupted text and a left-to-right autoregressive decoder. In principle, the pre-training procedure has two stages: (1) text is corrupted with an arbitrary noising function, and (2) a transformer-to-transformer model is learned to reconstruct the original text. 
Because BART has an autoregressive decoder, it can be
directly fine-tuned for most sequence generation tasks. Here, the encoder input is the a sequence of words, and the decoder generates outputs autoregressively, as shown in Figure \ref{figure:bart}. BART achieves new state-of-the art results on a number of text generation tasks, making it an ideal choice for generating similes. We refer the reader to \cite{lewis2019bart} for further details. 

For our task, we fine-tune BART by treating the literal input as encoder source and the simile as the the decoder target. Post fine-tuning at the inference step, we use top-k sampling strategy \cite{fan2018hierarchical} to generate similes conditioned on a test literal input.

\paragraph{Implementation details.}
Hyper-parameters, and essential details needed for reproducing experiments are given in Appendix A.1. 

\section{Experimental Setup}
To compare the quality of the generated similes, we benchmark SCOPE model  and human generations (HUMAN1 \& HUMAN2) described in Section \ref{section:evaldata} with three baseline systems described below
\subsection{Baseline Systems} 
Simile generation is a new task. The baselines outlined below have been used for other generation tasks. We adapt them to generate similes. 

 \begin{enumerate}
     \item \textbf{BART}: This is the pre-trained BART model. Since BART is a pre-trained sequence to sequence model, it can still be used for conditional text generation. To this end we use the same literal sentence (For example \textit{The city was beautiful}) as an input to the encoder and force the decoder to begin with same prefix by removing the adjective/adverb at the end and appending the comparator and the article (\textit{The city was like a}) and generate a simile.
     \item \textbf{Retrieval (RTRVL)}:  We also experiment with a retrieval approach where we retrieve a {\tt VEHICLE} from ConceptNet \cite{conceptnet} having the highest \textit{HasProperty} relation w.r.t our input (i.e., an adjective or adverb at the end of literal sentence) \footnote{ConceptNet is a weighted graph with multiple relations as can be viewed here http://conceptnet.io/ . We use `has property" for our work.There are multiple edges for objects with their properties. We choose edge with highest weight}. For the input \textit{The city was beautiful} we query ConceptNet with \textit{beautiful} and it returns \textit{sunset} as the {\tt VEHICLE} having highest weight for \textit{HasProperty beautiful}. We take this retrieved {\tt VEHICLE} and append it to the prefix ending in \textit{like a}. If the word is not in ConceptNet, we fall back to its synonyms obtained from Wordnet  \cite{miller1995wordnet}.
     \item \textbf{Metaphor Masking (META\_M)}: The third baseline is the metaphor generation model given a literal sentence described by \newcite{metagen2}. Following their approach, we fine-tune BART where we mask the adjective or adverb in the end of the literal sentence. The input is the masked text, with the hidden adjective or adverb (\textit{The city was \textbf{\textless MASK \textgreater}}), and the output is the original simile (\textit{The city was like a painting}). Through this learning paradigm, the model learns that it needs to generate simile when it encounters the mask token. At test time, we provide the model with the literal input, mask the adjective/adverb, and the model produces an output conditioned on the adjective/adverb masking training.
    
\end{enumerate}

\subsection{Evaluation Criteria}
\paragraph{Automatic evaluation.}
\textit{BLEU}~\cite{BLEU} is one of the most widely used automatic evaluation metric for generation tasks such as Machine Translation. However, for creative text generation, it is not ideal to expect significant n-gram overlaps between the machine-generated and the gold-standard sentences. We still report the BLEU scores for generated {\tt VEHICLE} after discarding the common prefix with the gold.

\textit{BERTScore}~\cite{zhang2019bertscore} has been used recently for evaluating text generation using contextualized embeddings and said to somewhat ameliorate the problems with BLEU. It computes a similarity score using contextual embeddings for each token in the candidate (here {\tt VEHICLE} in generated simile) with each token in the reference ({\tt VEHICLE} in human written simile).To compute F1-Score it uses Recall (matching each token in reference to a token in candidate) and Precision(matching each token in candidate to a token in reference).We report F1Score of \textit{BERTScore.}

\textit{Novelty.} To measure the model's generalization capability, we also want to test how well our models can generate novel content. We capture the proportion of generated {\tt VEHICLE} conditioned on an adverb/adjective literal {\tt PROPERTY} that does not appears in the training set.


\begin{table}[]
\small
\centering
\begin{tabular}{|l|l|l|l|l|}
\hline
       & \bf B-1 & \bf B-2 & \bf BERT-S & \bf NOVELTY \\ \hline
RTRVL   & 0.0    & 0.0    &   0.13 & 92.6         \\ \hline
BART    & 3.25  & 0.32 &       0.12 &  92.6    \\ \hline
META\_M & 3.73  & 0.96  &   0.15   & \textbf{93.3}    \\ \hline
SCOPE    & \textbf{8.03}  & \textbf{3.59}  & \textbf{0.18}       & 88.6    \\ \hline
\end{tabular}
\caption{Results using automatic metrics:  BLEU-1 (B-1), BLEU-2 (B-2), BERTScores (BERT-S) and Novelty. 
Boldface denotes the best results.}
\label{table:autoeval}
\end{table}

\begin{table}[]
\small
\centering
\begin{tabular}{|@{ }l@{ }|@{ }l@{}|@{ }l@{}|@{ }l@{}|@{ }l@{ }|}
\hline
\bf System & \bf C & \bf R1 & \bf R2 & \bf OQ \\ \hline
HUMAN1 & \textbf{3.61} (0.34)   & \textbf{3.74} (0.43)   & 3.90 (0.51)   & \textbf{3.54} (0.40)   \\ \hline
HUMAN2 & 3.46 (0.31)  & 3.72 (0.43)   & \textbf{3.97} (0.47)   &  3.44 (0.39)  \\ \hline\hline
RTRVL    & 1.90 (0.39)  &  1.85 (0.44)  & 1.73 (0.50)   & 1.85 (0.42)   \\ \hline
BART  & 2.68 (0.39)   & 2.78 (0.45)   & 2.75 (0.51)   &  2.61 (0.41)  \\ \hline
META\_M & 2.68 (0.42)  & 2.72 (0.46)   & 2.77 (0.47)   & 2.59 (0.41)   \\ \hline
SCOPE   & \underline{3.16} (0.35)  & \underline{3.50} (0.43)   & \underline{3.78} (0.52)   & \underline{3.32} (0.43)    \\ \hline
\end{tabular}
\caption{Human evaluation on several criteria of similes' quality for different systems' outputs and human written similes. 
We show average scores on a 1-5 scale with 1 denotes the worst and 5 be the best; the corresponding inter-annotator agreement (IAA) is in the parenthesis. Boldface denotes the best results and underscore denotes the second bests.}
\label{table:example3}
\end{table}

\begin{table}[t]
\small
\centering
\begin{tabular}{|p{0.3cm}|l|l|l|l|l|l|}
\hline
\multirow{2}{*}{} & \multicolumn{2}{l|}{\bf SCOPE/H1} & \multicolumn{2}{l|}{\bf SCOPE/H2} & \multicolumn{2}{l|}{\bf SCOPE/META\_M} \\ \cline{2-7} 
                  & w\%           & l\%           & w\%           & l\%           & w\%           & l\%           \\ \hline
C                 & 28.0                & \textbf{58.6}                  &  26.6               &    \textbf{57.3}              & \textbf{58.6}                &   31.3              \\ \hline
R1                &  37.3               &  \textbf{51.3}                &   33.3              &      \textbf{50.0}            &      \textbf{63.3}           &   18.0               \\ \hline
R2                &   42.6              &   \textbf{45.3}               &     37.3            &      \textbf{44.6}             &             \textbf{69.3}   &        17.3          \\ \hline
OQ                & 32.6                &    \textbf{54.6}              &   41.3              &       \textbf{50.0}           &         \textbf{68.6}        &   18.6               \\ \hline
\end{tabular}
\caption{Pairwise comparison between SCOPE and HUMAN1(H1), HUMAN2(H2), and META\_M. Win[w]\% (lose[l]\%) is the percentage of SCOPE gets a higher (lower) average score compared to HUMAN1, HUMAN2 and META\_M. The rest are ties.}
\label{table:example4}
\end{table}

\paragraph{Human evaluation.}
Automated metrics are not adequate on their own for evaluating methods to generate creative text so we present a human-based evaluation as well. We evaluate on a total of 900 utterances, 600 generated from 4 systems and 300 utterances generated by humans.  We proposed a set of 4 criteria to evaluate the generated output: (1) \textbf{Creativity (C)} (``How creative are the utterances?''), 
(2) \textbf{Overall Quality (OQ)} (``How good is the simile overall? ( \textit{Turk guidelines was to score based on how creative, well formed, meaningful and relevant it is with respect to the literal utterance})), (3) \textbf{Relevance1 (R1)} (``How relevant is the generated {\tt VEHICLE} in terms of portraying the {\tt PROPERTY}?'') and (4) \textbf{Relevance2 (R2)} (``How relevant is the {\tt VEHICLE} to the {\tt TOPIC} in the generation?''). As we evaluate on 4 separate dimensions for 900 utterances we have a total of 3600 evaluations. We hired Turkers on MTurk to rate outputs from the 4 systems and 2 humans. Each Turker was given the literal utterance as well as the 6 generate similes (randomly shuffled) 
Each criteria was rated on a scale from 1 (not at all) to 5 (very). Each utterance was rated by three separate Turkers. We hired 86,48,42,46 Turkers for the tasks of Creativity, Overall Quality, Relevance1, Relevance2 respectively. Turkers were paid at the rate of $20$ dollars an hour for the task. Further details in Appendix A.4 . 

\begin{table*}[!ht]
\small
\centering
\begin{tabular}{|@{ }p{2cm}@{ }|l@{ }|@{ }p{8.5cm}@{ }|@{ }l@{ }|@{ }l@{ }|@{ }l@{ }|@{ }l@{ }|}
\hline
Literal                                                                                                                                     & System & Simile                                                                                                                                  & R1 & R2 & C & OQ \\ \hline
\multirow{6}{*}{\begin{tabular}[c]{@{}p{2cm}@{}}It was obscene, but she was drawn to it, \textit{fascinated}\end{tabular}}  & HUMAN1    & \begin{tabular}[c]{@{}l@{}}    It was obscene, but she was drawn to it like a \textit{moth to a flame} \end{tabular}      & \textbf{5.0}  & \textbf{4.0}  &  3.0 & \textbf{4.7}   \\ 
                              \cline{2-7}
                              & HUMAN2    & \begin{tabular}[c]{@{}l@{}}    It was obscene, but she was drawn to it like it was\\ \textit{a bad boy in leather jacket} \end{tabular}       &  4.0 & 3.3  & \textbf{4.3}   &  1.7 \\ \cline{2-7} & RTRVL    & \begin{tabular}[c]{@{}l@{}}It was obscene, but she was drawn to it like a \textit{read}\end{tabular}       & 1.0  & 1.0  & 1.3   &1.3   \\
                              \cline{2-7} 
                                                             & BART    & \begin{tabular}[c]{@{}l@{}}It was obscene, but she was drawn to it like a \textit{magnet}\end{tabular}     &\textbf{5.0}   & \textbf{4.0}  & 2.7  & 2.7  \\ \cline{2-7} 
                                              & META\_M    & \begin{tabular}[c]{@{}l@{}}It was obscene, but she was drawn to it like a \textit{magnet}\end{tabular}       & \textbf{5.0}  & \textbf{4.0}  &2.7   & 2.7  \\                                                                                           \cline{2-7}
                                                            & SCOPE    & \begin{tabular}[c]{@{}l@{}} It was obscene, but she was drawn to it like a \textit{moth to a flame}\end{tabular} &  \textbf{5.0} & \textbf{4.0}  & 3.0  & \textbf{4.7}  \\ \cline{2-7}  \hline\hline
\multirow{6}{*}{\begin{tabular}[c]{@{}p{2cm}@{}}I start to prowl across the room \textit{warily} \end{tabular}}                      & HUMAN1    & \begin{tabular}[c]{@{}l@{}}   I start to prowl across the room like a \textit{ tightrope walker} \\ \textit{on dental floss} \end{tabular}      & 3.7  & \textbf{4.0}   & 2.7  & \textbf{ 5.0} \\ 
                              \cline{2-7}
                              & HUMAN2    & \begin{tabular}[c]{@{}l@{}}    I start to prowl across the room like a \textit{nervous criminal}\end{tabular}       & \textbf{4.7}   & 3.7  & 2.7  & 4.0  \\ \cline{2-7} & RTRVL    & \begin{tabular}[c]{@{}l@{}}I start to prowl across the room like a \textit{------}\end{tabular}       & 2.0   & 1.0  & 2.7  & 1.0  \\
                              \cline{2-7} 
                                           
                                                                                                                                            & BART    & \begin{tabular}[c]{@{}l@{}}I start to prowl across the room like a \textit{cat}\end{tabular}     & 2.7  & 3.3  & 3.7  & 3.3  \\ \cline{2-7} 
                                                                                        & META\_M    & \begin{tabular}[c]{@{}l@{}}I start to prowl across the room like a \textit{lion}\end{tabular}       & 2.7  & 3.7  & 3.3  & 2.7  \\                                                                                              \cline{2-7}
                                                                                        & SCOPE    & \begin{tabular}[c]{@{}l@{}} I start to prowl across the room like a \textit{cat stalking its prey}\end{tabular} & 3.0  & \textbf{4.0}  & \textbf{4.0}  & 4.0  \\ \cline{2-7}
                               \hline\hline
\multirow{6}{*}{\begin{tabular}[c]{@{}p{2cm}@{}}If it falls into the wrong hands it would be \textit{catastrophic}\end{tabular}}                                                   & HUMAN1    & \begin{tabular}[c]{@{}l@{}}If it falls into the wrong hands it would be like a \textit{nuclear apocalyse} \end{tabular}       &  4.0 & 4.3  & \textbf{4.7}  & \textbf{5.0}  \\
                              \cline{2-7}
                              & HUMAN2   & \begin{tabular}[c]{@{}l@{}}If it falls into the wrong hands it would be like \textit{World War III} \end{tabular}       &  \textbf{4.3} & \textbf{4.7}  & 4.0  & 4.7  \\ \cline{2-7} & RTRVL    & \begin{tabular}[c]{@{}l@{}}If it falls into the wrong hands it would be like a \textit{police officer}\end{tabular}       & 1.3 & 1.0  & 1.3  &1.0   \\
                              \cline{2-7}
                                                                                                                                            & BART    & \begin{tabular}[c]{@{}l@{}}If it falls into the wrong hands it would be like a \textit{gift to `terrorists'}\end{tabular}     &3.7   & 4.0  & 2.3  & 4.0  \\ \cline{2-7} 
                                                                             
                                                                                                                                            & META\_M    & \begin{tabular}[c]{@{}l@{}}If it falls into the wrong hands it would be like a \textit{gift}\end{tabular}       &   1.3 & 1.3  & 1.7  & 1.0  \\                                                                                              \cline{2-7}
& SCOPE    & \begin{tabular}[c]{@{}l@{}} If it falls into the wrong hands it would be like a \textit{nuclear bomb}\end{tabular} &\textbf{4.3}   &\textbf{4.7}   & 4.0  & \textbf{5.0}  \\ \cline{2-7} 
                               \hline\hline
                          
\multirow{6}{*}{\begin{tabular}[c]{@{}p{2cm}@{}}Having a \\thin figure,\\ he looked\\ \textit{unpleasant}\end{tabular}}   & HUMAN1    & \begin{tabular}[c]{@{}l@{}}   Having a thin figure, he looked like a  \textit{dry, overgrown blade}\\\textit{of grass} \end{tabular}      & 2.3  &3.7   & \textbf{4.7}  & \textbf{4.3}  \\ 
                              \cline{2-7}
                              & HUMAN2    & \begin{tabular}[c]{@{}l@{}}    Having a thin figure, he looked like a \textit{couch without cushions}\end{tabular}       & 2.0  & \textbf{4.7}  &  4.0 & 3.0  \\ \cline{2-7} & RTRVL    & \begin{tabular}[c]{@{}l@{}}Having a thin figure, he was looked like a \textit{ pain}\end{tabular}       & 2.3  &  1.0 & 2.0  & 1.3  \\
                              \cline{2-7} 
                                                                                                                                            & BART    & \begin{tabular}[c]{@{}l@{}}Having a thin figure, he looked like a \textit{man}\end{tabular}     & 2.3  & 1.0  & 1.0  &1.7   \\ \cline{2-7} 
                                                                                                                                            
                                                                                                                                            & META\_M    & \begin{tabular}[c]{@{}l@{}}Having a thin figure, he looked like a \textit{child}\end{tabular}       & 2.0  & 2.3  &  1.3 & 2.7  \\                                                                                              \cline{2-7}
& SCOPE    & \begin{tabular}[c]{@{}l@{}} Having a thin figure, he looked like a \textit{walking corpse}\end{tabular} & \textbf{4.0}   & 4.3  & 3.0  & 3.7  \\ \cline{2-7} \hline\hline
\multirow{6}{*}{\begin{tabular}[c]{@{}p{2cm}@{}}Keaton was lying face down on the ground \textit{lamely}\end{tabular}}   & HUMAN1    & \begin{tabular}[c]{@{}l@{}}   Keaton was lying face down on the ground like a \textit{snapped twig} \end{tabular}      & 1.3  & 4.0   & 3.7  & 2.3  \\ 
                              \cline{2-7}
                              & HUMAN2    & \begin{tabular}[c]{@{}l@{}}    Keaton was lying face down on the ground like a \textit{maimed animal}\end{tabular}       & 3.7  & \textbf{4.3}  &  \textbf{4.0} & 3.3  \\ \cline{2-7} & RTRVL    & \begin{tabular}[c]{@{}l@{}}Keaton was lying face down on the ground like a \textit{life}\end{tabular}       & 1.3  &  1.0 & 2.3  & 1.0  \\
                              \cline{2-7} 
                                                                                                                                            & BART    & \begin{tabular}[c]{@{}l@{}}Keaton was lying face down on the ground like a \textit{ zombie}\end{tabular}     & 3.3  & 2.7  & 3.0  &3.0   \\ \cline{2-7} 
                                                                                                                                            
                                                                                                                                            & META\_M    & \begin{tabular}[c]{@{}l@{}}Keaton was lying face down on the ground like a \textit{sack of potatoes}\end{tabular}       & \textbf{4.0}  & 3.7  &  2.3 & 3.3  \\                                                                                              \cline{2-7}
& SCOPE    & \begin{tabular}[c]{@{}l@{}} Keaton was lying face down on the ground like a \textit{limp corpse}\end{tabular} & \textbf{4.0}   & 4.0  & 3.7  & \textbf{4.3}  \\
\cline{2-7} \hline\hline

\multirow{6}{*}{\begin{tabular}[c]{@{}p{2cm}@{}}As a human tick hybrid, life can sometimes be \textit{tricky}\end{tabular}}   & HUMAN1    & \begin{tabular}[c]{@{}l@{}}   As a human tick hybrid, life can sometimes be like \\  \textit{sucking the nectar out of dried fruit} \end{tabular}      & 2.3  &3.7   & \textbf{4.7}  & 3.3  \\ 
                              \cline{2-7}
                              & HUMAN2    & \begin{tabular}[c]{@{}l@{}}     As a human tick hybrid, life can sometimes be like \\  \textit{interspecies balancing act}\end{tabular}       & \textbf{4.3}  & 4.3  &  3.0 & 4.0 \\ \cline{2-7} & RTRVL    & \begin{tabular}[c]{@{}l@{}} As a human tick hybrid, life can sometimes be like a  \textit{ceiling}\end{tabular}       & 1.3  &  1.0 & 1.3  & 1.7  \\
                              \cline{2-7} 
                                                                                                                                            & BART    & \begin{tabular}[c]{@{}l@{}} As a human tick hybrid, life can sometimes be like a  \textit{zoo}\end{tabular}     & 2.3  & 2.7  & 2.3  &2.7   \\ \cline{2-7} 
                                                                                                                                            
                                                                                                                                            & META\_M    & \begin{tabular}[c]{@{}l@{}} As a human tick hybrid, life can sometimes be like \textit{dream}\end{tabular}       & 1.3  & 2.3  &  2.0 & 2.0  \\                                                                                              \cline{2-7}
& SCOPE    & \begin{tabular}[c]{@{}l@{}}  As a human tick hybrid, life can sometimes be like a  \textit{slippery slope}\end{tabular} & \textbf{4.3}   & \textbf{4.7}  & 2.7  & \textbf{4.3}  \\ \cline{2-7} \hline
                              
\end{tabular}
\caption{Examples of generated outputs from different systems (with human written similes as references). 
We show average scores (over three annotators) on a 1-5 scale with 1 denotes the worst and 5 be the best. The italics texts in the literal column represent the PROPERTY while those in Simile column represents the generated VEHICLE. Boldface indicates the best results. More examples in Appendix A.3}
\label{table:example5}
\end{table*}

\section{Experimental Results} \label{section:results}
\subsection{Automatic Evaluation}
Table \ref{table:autoeval} shows BLEU-1, BLEU-2 and BERTScore of our system compared to the three baselines. The low scores can be attributed to the nature of creative NLG tasks.
To further validate this we also compute the BLEU-1 and BLEU-2 score between the two literary experts treating one as reference and other as candidate and get scores of $4.12$ and $0.52$ respectively. BERTScore is often a better metric as it utilizes contextualized embeddings. For example for a candidate [\textbf{desert}] with multi-reference as [[\textbf{sandy death trap}],[\textbf{wasteland}]] , we get a BERTscore of 0.99 while 
BLEU score is 0.0.
Finally our best model SCOPE emerges as the winner for both BLEU and BERTScore. For novelty SCOPE can still generate novel content 88\% of the time proving it is generalizable to unseen test data. Further there are 5558 unique {\tt PROPERTY} in training data and 41\% of {\tt PROPERTY} in testing data does not appear in training, showing our model is generalizable on unseen {\tt PROPERTY} as well.

\subsection{Human Evaluation Scores}
Table ~\ref{table:example3} presents the scores of the aforementioned evaluation criteria for our model and the baselines on the test set. 
The results 
show that SCOPE is significantly ($p<.001$ according to approximate randomization test) better than the baselines on all four criteria.
For all metrics our best system is comparable to humans. We also computed Pearson's correlation between OQ with other metrics and observed that R1 and R2 had moderate correlation of 0.54 and 0.52 with OQ , while C was fairly correlated (0.31) to OQ suggesting a relevance matters when deciding the quality of a simile.

\paragraph{Pairwise Comparison between systems.}
Table \ref{table:example4} shows the pairwise comparisons between the SCOPE and human generated simile (HUMAN1 and HUMAN2), and META\_M \cite{metagen2}, respectively. Given a pair of inputs, we decide win/lose/tie by comparing the average scores (over three Turkers) of both outputs. We see that SCOPE outperforms META\_M on all the metrics. For overall quality, although it is a given that literary experts are better, the SCOPE model still has a winning rate of 32.6\% and 41.3\% respectively.

\begin{figure}
\centering
\includegraphics[scale=0.17]{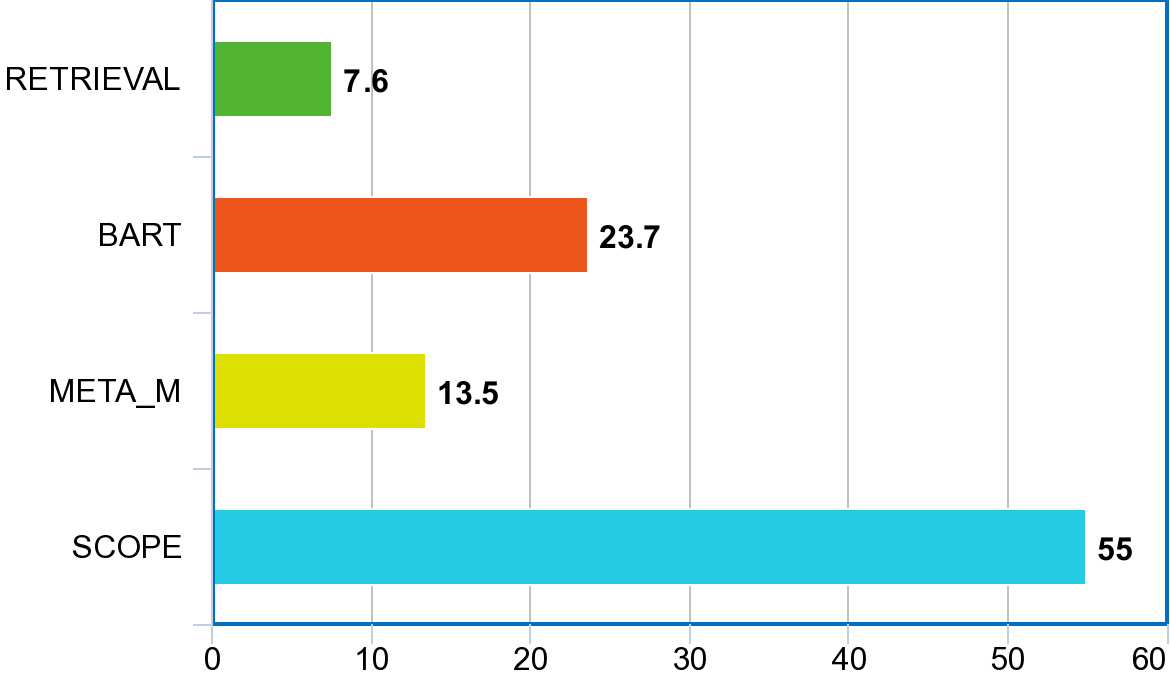}
\caption{\label{figure:cn} Barchart showing the percent of times each individual system won in terms of Overall Quality.}
\end{figure}

\begin{table}[]
\small
\centering
\begin{tabular}{|l|}
\hline
\begin{tabular}[c]{@{}l@{}} \textbf{Storyline}: sky $\rightarrow$ sunset $\rightarrow$ walk \\
$\rightarrow$ walked $\rightarrow$ beautiful
\end{tabular} \\ \hline
\begin{tabular}[c]{@{}l@{}}The sky was \sout{beautiful} [\textit{like a blue canvas}]. Jane \\wanted to see the sunset. She decided to go for a \\walk.She walked for a long time.When she \\was done she saw the sunset was beautiful.\end{tabular} \\ \hline
\begin{tabular}[c]{@{}l@{}} \textbf{Title}: car accident \textbf{Storyline}: driving $\rightarrow$ hit \\ $\rightarrow$ hit 
$\rightarrow$ car $\rightarrow$ fixed
\end{tabular} \\ \hline
\begin{tabular}[c]{@{}l@{}}Tom was driving down the road. Suddenly he\\ hit a tree. He swerved and hit a pole. Tom’s car\\ was \sout{totaled} \textit{[like a wreck]}. Luckily he was\\ able to get it fixed\end{tabular} \\ \hline
\end{tabular}
\caption{\label{table:story}An example of a GPT-2 generated short stories with the title \textbf{Sunset} and \textbf{Car Accident}. We replace the  literal sentences with generated similes from SCOPE.}
\label{table:example7}
\end{table}

\section{Qualitative Analysis}
Table \ref{table:example5} demonstrates several generation outputs from different systems along with human judgements on individual criteria. We observe that often our model is better than at least one human on a certain criteria while outperforming the baselines by a large margin.

\subsection{Role of Relevance}
While conditioning on the context of literal sentences might lead to grammatically correct similes, they are often not meaningful and relevant to the {\tt PROPERTY} in consideration. META\_M generates similes by fine-tuning BART on literal sentences where the common sense {\tt PROPERTY} is masked. The lack of relevance mapping during fine-tuning often leads to improper generations. For instance, referring to Table \ref{table:example5}, the context of `falling into the wrong hands' is more likely to lead to something bad and hence here `gift' is not appropriate while `nuclear bomb' is. One possible way of incorporating relevance is through common sense knowledge.

\subsection{Role of Context}
The role of context is necessary for simile generation. For example given the literal input \textit{`But times are hard, and silver bullets are \textbf{expensive}'} even though ConceptNet tells us \textbf{diamonds} are objects with \textit{HasProperty} expensive, a generated simile by RTRVL model \textit{`But times are hard, and silver bullets are  like a \textbf{diamond}'} seems inappropriate suggesting that a context leads to better generation. Our SCOPE model generates \textit{`But times are hard, and silver bullets are  like a \textbf{luxury item}'}


\section{Task-based Evaluation: Simile for Story Generation} \label{section:story}
Similes are often used to evoke imagery. Generating or transforming text to be evocative can be useful for computational journalism \cite{spanghermodeling}, poetry generation \cite{ghazvininejad-etal-2017-hafez,van2020automatic} and story writing \cite{goldfarb2020content,yao2019plan}. Table \ref{table:example7} shows how we can use our simile generation module as a post processing step to replace literal sentences leading to more expressive and creative stories. To further test this hypothesis we conduct an experiment further outlined below.

\begin{table}[]
\small
\centering
\begin{tabular}{|c|c|c|}
\hline
GPT2 & GPT2+META\_M & GPT2+SCOPE \\ \hline
23\%   & 25\%       & \textbf{42\% }     \\ \hline
\end{tabular}
\caption{\label{tab:analysis}Win\% (in terms of average score over three annotators) of stories generated with only GPT2, GPT2 with META\_M or SCOPE simile post processing. The rest are ties.}
\end{table}

\subsection{Story Generation} We use the ROCStories \cite{mostafazadeh2016corpus} dataset to generate stories using the \textit{Plan and Write } model outlined by \citet{yao2019plan}. We introduce a two step pipeline procedure where we fine-tune a pre-trained GPT2 \cite{gpt} model on  titles and storyline from the training set to generate a storyline given a title (Row 1 Table \ref{table:story}). In parallel, we also fine-tune GPT2 on storylines and stories from the training set to generate a story given a storyline (Row 2 Table \ref{table:story}). At test time, we generate a storyline using an input title first and then use the generated storyline to generate a story.

\subsection{Post Processing} There can be multiple sentences ending with an adjective or adverb and replacing each of them with a simile might lead to over-embellishment. Under such situations we feed only one randomly selected sentence to SCOPE and META\_M module and replace the sentence in GPT2 generated story with the output from SCOPE or META\_M, respectively.

\subsection{Human evaluation.} We randomly select 50 titles from ROCStories data set and generate stories as described above. We postprocess it using both SCOPE and META\_M separately. Thus for each title we have 3 stories 1) the original GPT2 story 2)the GPT2 story postprocessed with SCOPE 3)the GPT2 story postprocessed with META\_M. For each given titles, we present these 3 stories each to workers in AMT and ask them to score them in a range of 1(poor) to 5 (excellent) based on creativity and evocativeness. Experimental results from Table \ref{tab:analysis} prove that effective usage of similes can improve evocativeness and reception of machine generated stories.



\section{Related Work}

Simile generation is a relatively new task. Most prior work
has focused on detection of similes. The closest task in NLP to simile generation is generating metaphors. However it should be noted the overlap between the expressive range of similes and metaphors is known to be only partial: there
are similes that cannot be rephrased as metaphors, similarly the other way around~\cite{israel2004simile}.

\subsection{Simile Detection and Analysis}
\citet{simile1} proposed frameworks for annotating similes from product reviews by considering their semantic and syntactic characteristics as well as the challenges inherent to the automatic detection of similes. \citet{simile3,simile4} built computational models to recognize affective polarity and implicit properties in similes. Unlike these works, we focus on generating similes by transforming a literal sentence while still being faithful to the property in context.



\subsection{Metaphor Generation}

Earlier works in metaphor generation \cite{abe2006computational,terai2010computational} were conducted on a  lexical or phrase level, using template and heuristic-based methods. \cite{metaphoria} presented an interactive system for collaboratively
writing metaphors with a computer. They use an open source knowledge graph
and a modified Word Mover’s Distance algorithm to find
a large, ranked list of suggested metaphorical connections. Word embedding approaches \cite{gagliano2016intersecting} have also been used for metaphor generation. \cite{young1987metaphor} also present a relational data base method for automatic metaphor generation.  However, the metaphors generated through these methods do not take semantic context into consideration and lack the flexibility and creativity necessary to instantiate similes through a natural language sentence. 



\citet{metagen1} use neural models to generate metaphoric expressions given a literal input in an unsupervised manner. \citet{metagen2} develop a new framework dubbed `metaphor masking' where they train a supervised seq2seq model with input as the masked text, where they mask or hide the metaphorical verb while preserving the original text as the output. However, both these works hinge on metaphoric verbs unlike similes where we not only need to replace the literal property with a vehicle but it also needs to be relevant to the context and the tenor. Additionally we also use \cite{metagen2} as a baseline and show that their approach may not be the best way for generating similes.

\section{Conclusion}
We establish a new task for NLG: simile generation from literal sentences. We propose a novel way of creating parallel corpora and a transfer-learning approach for generating similes. Human and automatic evaluations show that our best model is successful at generating similes. Our experimental results further show that to truly be able to generate similes based on actual metaphoric or conceptual mappings, it is important to incorporate some common sense knowledge about the topics and their properties. Future directions include exploration of other knowledge bases to help the inference and applying our simile generation approach to different creative NLG tasks such as sarcasm \cite{chakrabarty-etal-2020-r}, hyperbole \cite{hyperbole} etc.

\section*{Acknowledgments}

This work was supported by the CwC program under Contract W911NF-15-1-0543 with the US Defense Advanced Research Projects Agency (DARPA). The views expressed are those of the authors and do not reflect the official policy or position of the Department of Defense or the U.S. Government.
The authors would like to thank Kai-Wei Chang, Christopher Hidey, Christopher Robert Kedzie, Anusha Bala and Liunian Harold Li for useful discussions. The authors also thank members of PLUSLab at the University Of California Los Angeles and University Of Southern California and the anonymous reviewers for helpful comments. The first author also wants to acknowledge his father Tridib Chakrabarty who bravely fought through this pandemic and left us for the heavenly abode through the period of writing this paper.



\bibliography{anthology,emnlp2020}
\bibliographystyle{acl_natbib}

\appendix

\section{Appendix}
\subsection{Hyper-Parameters and Other Experimental Settings}

For retrieving commonsense properties of the vehicle, we use the pre-trained COMET model \footnote{\url{https://github.com/atcbosselut/comet-commonsense}} and retrieve top 5 candidates for each input.
\begin{enumerate}
    \item{\textbf{No of Parameters:}} For BART we use the BART large checkpoint (400M parameters) and use the implementation by FAIRSEQ \cite{ott2019fairseq}
     \footnote{\url{https://github.com/pytorch/fairseq/tree/master/examples/bart}}
    \item{\textbf{No of Epochs:}} We fine-tune pre-trained BART for 17 epochs for SCOPE model.
    \item{\textbf{Training Time:}} Our training time is 52 minutes
    \item{\textbf{Hardware Configuration:}} We use 4 RTX 2080 GPU
    \item{\textbf{Training Hyper parameters:}} We use the same parameters mentioned in the github repo where BART was fine-tuned for CNN-DM summarization task with the exception of MAX-TOKENS (size of each mini-batch, in terms of the number of tokens.) being 1024 for us
    \item{\textbf{Decoding Strategy \& Hyper Parameters:}} For decoding we  generate similes from our models using a top-k random sampling scheme \cite{fan2018hierarchical}. At each timestep, the model generates the probability of each word in the vocabulary being the likely next word. We randomly sample from the k = 5 most likely candidates from this distribution. We also use a softmax temperature of 0.7.

\end{enumerate}

\begin{table}[]
\small
\centering
\begin{tabular}{|l|l|l|}
\hline
                                                             & \#WORKERS & $\alpha$ \\ \hline
C                                                  & 86        & 0.36                  \\ \hline
\begin{tabular}[c]{@{}l@{}}OP\end{tabular} & 48        & 0.41                  \\ \hline
R1                                                   & 42        & 0.44                  \\ \hline
R2                                                   & 46        & 0.49                  \\ \hline
\end{tabular}
\caption{\label{tab:1} C, R1,R2 and OQ denote Creativity, Relevance of Vehicle w.r.t Property, Relevance of Tenor to Vehicle and Overall Quality. WORKERS  denote number of workers employed for each task and $\alpha$ denotes Krippendorff's alpha ($\alpha$ ) , reliability coefficient used for our study}
\vspace{-3ex}
\end{table}

\subsection{Dataset Assumptions}

\begin{figure}[ht]
\centering
\includegraphics[scale=0.16]{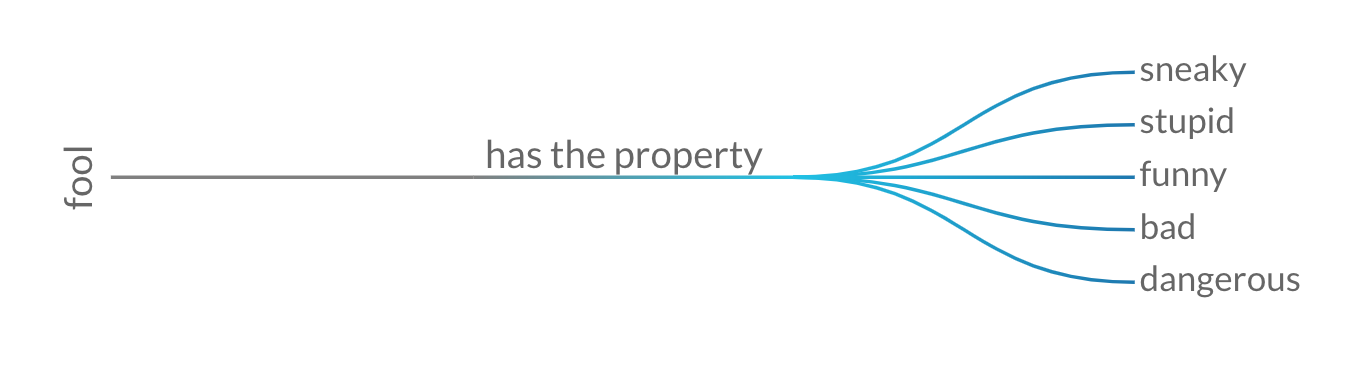}
\caption{\label{figure:comet} Proprty associated with fool} 
\vspace{-1em}
\end{figure}

While distant supervision is often used to collect a lot of data without human/ expert annotation through this process we introduce, noise in our self labeled similes. For example the sentence \textit{I feel like a fool} is ideally not a simile.We notice 1.1\% of the training data with PNP in {\tt TOPIC} and typically $<=6$ in token count such as \textit{I would like a , I don't like a, I feel like a, I think like a}. However our transformation method still works here. Based on Figure \ref{figure:comet} we see the common sense properties associated for \textit{fool} are \textit{sneaky, stupid,funny,dangerous, bad}. Our best literal transformation for \textit{I feel like a fool} is then \textit{I feel stupid}. So even though there is some noise this method still benefits our training procedure


\begin{table*}[!ht]
\small
\centering
\begin{tabular}{|@{ }p{2cm}@{ }|l@{ }|@{ }p{8.5cm}@{ }|@{ }l@{ }|@{ }l@{ }|@{ }l@{ }|@{ }l@{ }|}
\hline
Literal                                                                                                                                     & System & Simile                                                                                                                                  & R1 & R2 & C & OQ \\ \hline
\multirow{6}{*}{\begin{tabular}[c]{@{}p{2cm}@{}}From the day you were born, you've been \textit{invincible}\end{tabular}}  & HUMAN1    & \begin{tabular}[c]{@{}l@{}}    From the day you were born, you've been \\like a \textit{well-seasoned superhero} \end{tabular}      & 4.0  & 4.0  &  4.0 & 3.3   \\ 
                              \cline{2-7}
                              & HUMAN2    & \begin{tabular}[c]{@{}l@{}}   From the day you were born, you've been like \textit{Superman
                              }\end{tabular}       &  \textbf{4.7} & 4.7  & \textbf{4.3}   &  1.7 \\ \cline{2-7} & RTRVL    & \begin{tabular}[c]{@{}l@{}}From the day you were born, you've been like a \textit{-------}\end{tabular}       & 1.0  & 4.0  & 1.0   &1.0   \\
                              \cline{2-7} 
                                                             & BART    & \begin{tabular}[c]{@{}l@{}}From the day you were born, you've been like a \textit{son}\end{tabular}     &\textbf{5.0}   & 1.3  & 1.0  & 1.3  \\ \cline{2-7} 
                                              & META\_M    & \begin{tabular}[c]{@{}l@{}}From the day you were born, you've been like a \textit{son to me}\end{tabular}       & 1.0 & 1.0  &1.7   & 1.3  \\                                                                                           \cline{2-7}
                                                            & SCOPE    & \begin{tabular}[c]{@{}l@{}}From the day you were born, you've been like a \textit{superhero}\end{tabular} &  \textbf{4.7} & \textbf{5.0}  & 4.0  & \textbf{4.3}  \\ \cline{2-7}  \hline\hline
\multirow{6}{*}{\begin{tabular}[c]{@{}p{2cm}@{}}For centuries, the Tyrant has made life \textit{miserable} \end{tabular}}                      & HUMAN1    & \begin{tabular}[c]{@{}l@{}}   For centuries, the Tyrant has made life like an\\ \textit{ impatient storm reaching to be a hurricane} \end{tabular}      & 2.7  & 2.7   & 4.0  & 3.3 \\ 
                              \cline{2-7}
                              & HUMAN2    & \begin{tabular}[c]{@{}l@{}}  For centuries, the Tyrant has made life like a \textit{dreary prison}\end{tabular}       & \textbf{5.0}   & 3.7  & \textbf{4.3}  & 3.7  \\ \cline{2-7} & RTRVL    & \begin{tabular}[c]{@{}l@{}}For centuries, the Tyrant has made life like  a \textit{------}\end{tabular}       & 1.3   & 1.0  & 1.3  & 1.7  \\
                              \cline{2-7} 
                                           
                                                                                                                                            & BART    & \begin{tabular}[c]{@{}l@{}}For centuries, the Tyrant has made life like a \textit{prison}\end{tabular}     & 4.3  & 3.0  & 3.3  & 3.3  \\ \cline{2-7} 
                                                                                        & META\_M    & \begin{tabular}[c]{@{}l@{}}For centuries, the Tyrant has made life like a \textit{prison in this country}\end{tabular}       & 4.3  & 3.0  & 3.3  & 3.7  \\                                                                                              \cline{2-7}
                                                                                        & SCOPE    & \begin{tabular}[c]{@{}l@{}} For centuries, the Tyrant has made life like a \textit{living hell}\end{tabular} & 4.7  & \textbf{5.0}  & 4.0  & \textbf{4.0}  \\ \cline{2-7}
                               \hline\hline
\multirow{6}{*}{\begin{tabular}[c]{@{}p{2cm}@{}}Adrenaline shot through him \textit{powerful}\end{tabular}}                                                   & HUMAN1    & \begin{tabular}[c]{@{}l@{}}Adrenaline shot through him like a \textit{lightning bolt} \end{tabular}       &  3.7 & \textbf{4.3}  & 3.3  & \textbf{4.7}  \\
                              \cline{2-7}
                              & HUMAN2   & \begin{tabular}[c]{@{}l@{}}Adrenaline shot through him like a \textit{hypodermic injection} \end{tabular}       &  3.7 & 3.7  & 3.0  & 2.7  \\ \cline{2-7} & RTRVL    & \begin{tabular}[c]{@{}l@{}}Adrenaline shot through him like a \textit{natural energy}\end{tabular}       & 2.7 & 2.3  & 2.7  &2.7  \\
                              \cline{2-7}
                                                                                                                                            & BART    & \begin{tabular}[c]{@{}l@{}}Adrenaline shot through him like a \textit{bullet}\end{tabular}     &\textbf{4.3}   & \textbf{4.3}  & \textbf{4.3}  & 3.7  \\ \cline{2-7} 
                                                                             
                                                                                                                                            & META\_M    & \begin{tabular}[c]{@{}l@{}}Adrenaline shot through him like a \textit{bullet}\end{tabular}       &   \textbf{4.3} & \textbf{4.3}  & \textbf{4.3}  & 3.7  \\                                                                                              \cline{2-7}
& SCOPE    & \begin{tabular}[c]{@{}l@{}} Adrenaline shot through him like a \textit{bolt of lightning}\end{tabular} & 3.3   &\textbf{4.3}   & 4.0  & \textbf{4.7}  \\ \cline{2-7} 
                               \hline\hline
                          
 \cline{2-7} \hline\hline
\multirow{6}{*}{\begin{tabular}[c]{@{}p{2cm}@{}}Constructing the flat pack TV cabinet was meant to be \textit{easy}\end{tabular}}  & HUMAN1    & \begin{tabular}[c]{@{}l@{}}    Constructing the flat pack TV cabinet was meant to be like\\ \textit{putting on velcro shoes} \end{tabular}      & 4.0  & 4.7  &  \textbf{5.0} & 3.7   \\ 
                              \cline{2-7}
                              & HUMAN2    & \begin{tabular}[c]{@{}l@{}}   Constructing the flat pack TV cabinet was meant to be like\\ \textit{turning on a light} \end{tabular}       &  \textbf{4.3} & 4.0  & 3.0   &  3.0 \\ \cline{2-7} & RTRVL    & \begin{tabular}[c]{@{}l@{}}Constructing the flat pack TV cabinet was meant to be like a\\ \textit{learn to change car tire}\end{tabular}       & 2.0  & 2.3  & 2.3   &2.7   \\
                              \cline{2-7} 
                                                             & BART    & \begin{tabular}[c]{@{}l@{}}Constructing the flat pack TV cabinet was meant to be like a\\ \textit{Lego set}\end{tabular}     &3.0   & 2.3  & 2.7  & 4.7  \\ \cline{2-7} 
                                              & META\_M    & \begin{tabular}[c]{@{}l@{}}Constructing the flat pack TV cabinet was meant to be like a \textit{house}\end{tabular}       & 1.0  & 1.0  &1.3   & 2.3  \\                                                                                           \cline{2-7}
                                                            & SCOPE    & \begin{tabular}[c]{@{}l@{}} Constructing the flat pack TV cabinet was meant to be like a\\ \textit{cake walk}\end{tabular} &  \textbf{5.0} & \textbf{4.3}  & 3.0  & \textbf{4.3}  \\ \cline{2-7}  \hline\hline
\multirow{6}{*}{\begin{tabular}[c]{@{}p{2cm}@{}}You are an oracle whose predictions have always come  \textit{true}\end{tabular}}                                                   & HUMAN1    & \begin{tabular}[c]{@{}l@{}}You are an oracle whose predictions have always come true like\\ the \textit{rising sun} \end{tabular}       &  3.7 & \textbf{4.3}  & \textbf{4.0}  & 2.7  \\
                              \cline{2-7}
                              & HUMAN2   & \begin{tabular}[c]{@{}l@{}}You are an oracle whose predictions have always come true like\\ \textit{highly researched hypotheses} \end{tabular}       &  3.0 & 4.0  & 1.7  & 2.7  \\ \cline{2-7} & RTRVL    & \begin{tabular}[c]{@{}l@{}}You are an oracle whose predictions have always come true like a\\ \textit{fact}\end{tabular}       & 3.7 & \textbf{4.3} & 2.0  &3.0   \\
                              \cline{2-7}
                                                                                                                                            & BART    & \begin{tabular}[c]{@{}l@{}}You are an oracle whose predictions have always come true like a\\ \textit{man of action}\end{tabular}     &2.0   & 2.0  & 2.7  & 2.0  \\ \cline{2-7} 
                                                                             
                                                                                                                                            & META\_M    & \begin{tabular}[c]{@{}l@{}}You are an oracle whose predictions have always come true like a\\ \textit{bolt from the blue}\end{tabular}       &   2.7 & 2.3  & 3.3  & 3.0  \\                                                                                              \cline{2-7}
& SCOPE    & \begin{tabular}[c]{@{}l@{}} You are an oracle whose predictions have always come true like a\\ \textit{prophecy}\end{tabular} &3.0   & 2.7   & 3.7  & \textbf{4.0}  \\ \cline{2-7} 
                               \hline
                              
\end{tabular}
\caption{\label{tab:example}Examples of generated outputs from different systems (with human written similes as references).We show average scores (over three annotators) on a 1-5 scale where 1 denotes the worst and 5 be the best. The italics texts in the literal column represent the PROPERTY while those in Simile column represents the generated VEHICLE. Boldface indicates the best results.}
\label{table:example5}
\end{table*}

\begin{figure*}[!ht]
\centering
\includegraphics[width=\textwidth,frame]{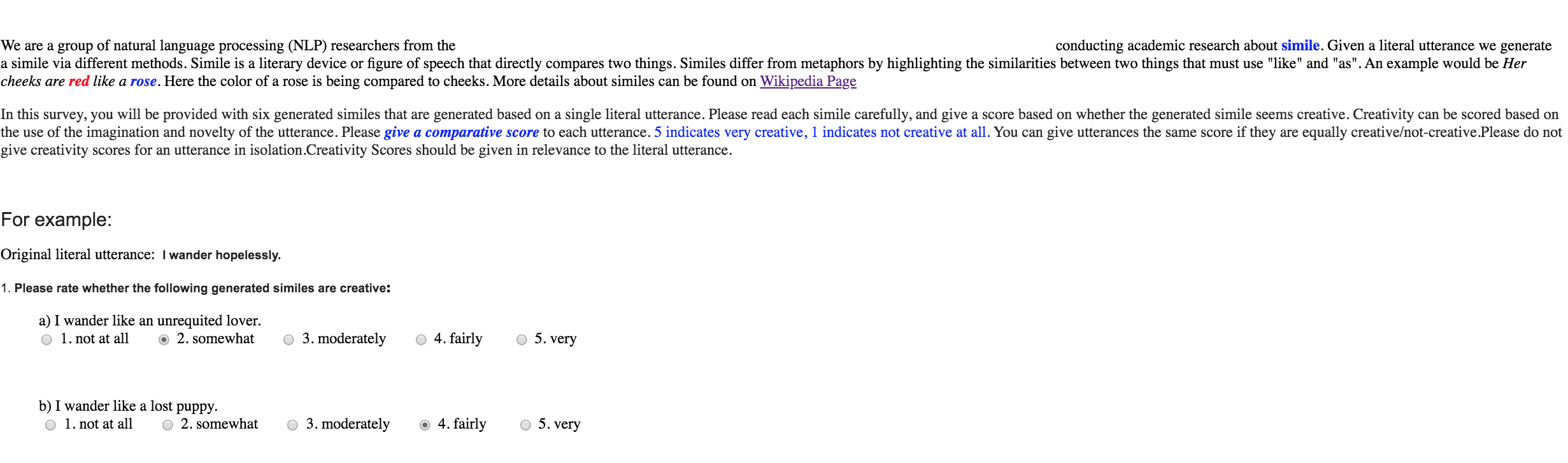}
\caption{\label{figure:C} MTurk interface for scoring \textbf{Creativity}}
\end{figure*}
\begin{figure*}[!ht]
\centering
\includegraphics[width=\textwidth,frame]{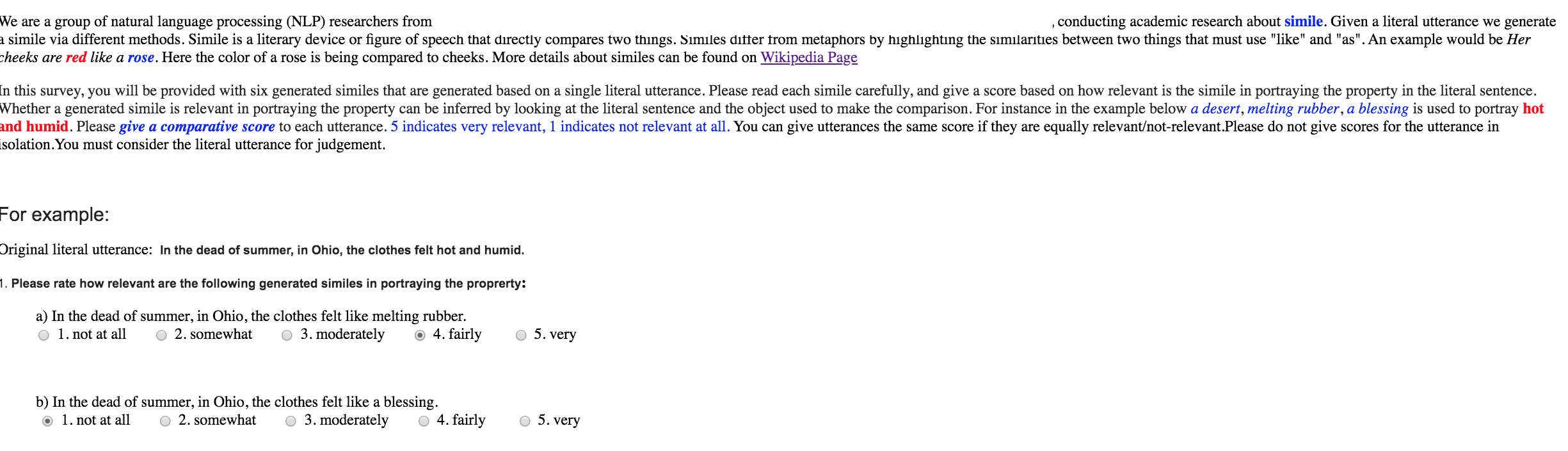}
\caption{\label{figure:R1} MTurk interface for scoring \textbf{Relevance1}}
\end{figure*}

\begin{figure*}[!ht]
\centering
\includegraphics[width=\textwidth,frame]{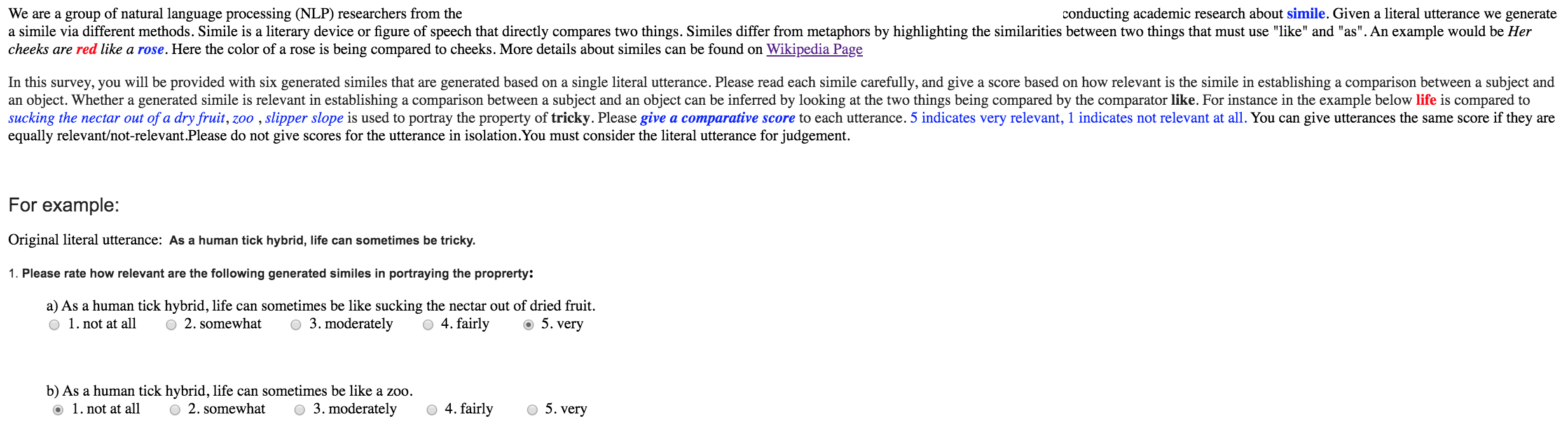}
\caption{\label{figure:R2} MTurk interface for scoring \textbf{Relevance2}}
\end{figure*}

\begin{figure*}[!ht]
\centering
\includegraphics[width=\textwidth,frame]{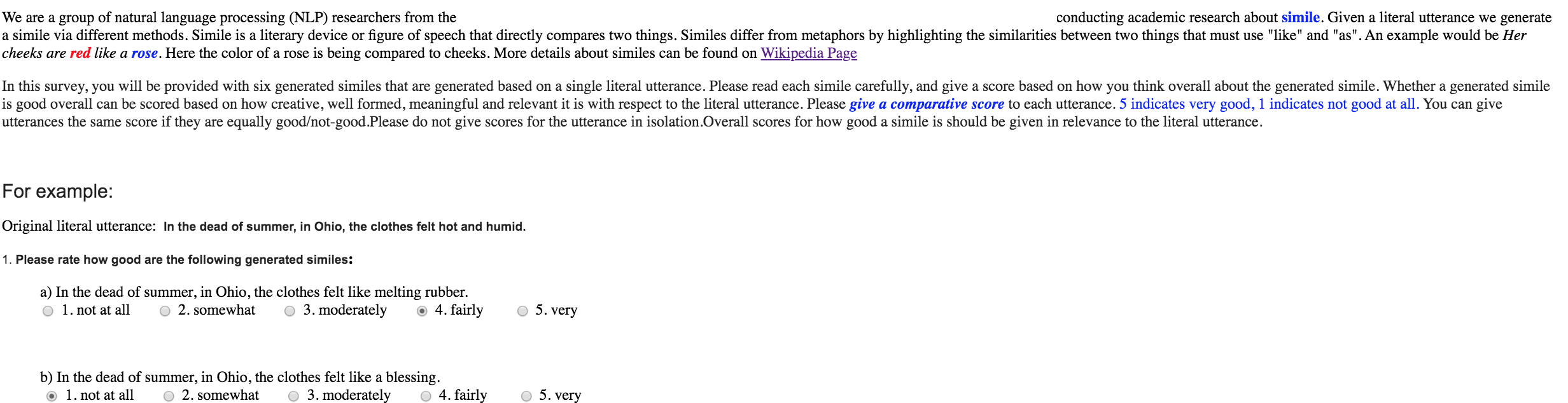}
\caption{\label{figure:OQ} MTurk interface for scoring \textbf{Overall Quality}}
\vspace{-1em}
\end{figure*}

\subsection{Examples}

Table \ref{tab:example} shows generations from all 4 systems along with gold similes and how turkers scored them on a scale of 1 to 5 for C,R1,R2 and OQ.

\subsection{Amazon Mechanical Turk Settings}

The 2nd column Table \ref{tab:1} shows the number of distinct workers employed for each task. Column 3 shows inter-rater agreement between workers. Except for Creativity, for the other 3 tasks workers are moderately correlated. For creativity workers are fairly correlated.

Figure \ref{figure:C},\ref{figure:R1},\ref{figure:R2} and \ref{figure:OQ} show the Amazon Mechanical Turk interfaces for the tasks of \textit{Creativity (C)} (``How creative are the utterances ?''), (2) \textit{Relevance1 (R1)} (``How relevant is the generated {\tt VEHICLE} in terms of portraying the {\tt PROPERTY}?'') and (3) \textit{Relevance2 (R2)} (``How relevant is the {\tt VEHICLE} to the {\tt TOPIC} in the generation?'')
(4) \textit{Overall Quality (OQ)} (``How good is the simile overall ?''. As can be seen we provide with explicit examples and a clear description of the task to turkers. We also mention and highlight the importance of evaluating similes along with input and not in isolation.

\subsection{GPT2 generated stories preprocessed with SCOPE}
Table \ref{table:story} shows several example stories where a literal sentence has been replaced by a simile.
\begin{table}[!ht]
\centering
\small
\begin{tabular}{|l|}
\hline
\begin{tabular}[c]{@{}l@{}} \textbf{Title}: a gift from the mentor  \textbf{Storyline}: loved $\rightarrow$ \\playing $\rightarrow$ promised 
$\rightarrow$ tried $\rightarrow$ surprised
\end{tabular} \\ \hline
\begin{tabular}[c]{@{}l@{}}Harry loved playing tennis. One day while \\playing he broke his racket. His coach had\\ promised to buy him a new racket if he\\ practiced. Harry tried hard to practice and was \\confident in his new racket. To his surprise  his\\ coach bought him a racket for his birthday and \\he was \sout{ecstatic } \textit{[like a child on Christmas day]} \end{tabular} \\ \hline
\hline
\begin{tabular}[c]{@{}l@{}} \textbf{Title}: The pet bug \textbf{Storyline}: playing $\rightarrow$ caught\\ $\rightarrow$ bug
$\rightarrow$ hoped $\rightarrow$ release
\end{tabular} \\ \hline
\begin{tabular}[c]{@{}l@{}}Oliver was playing in his yard. Suddenly he \\spotted a bug he hadn't caught. The bug was a\\ big beetle. He hoped it would be there \sout{forever} \\ \textit{[like a shadow]}. But unfortunately it was too\\ late to release it\end{tabular} \\ \hline
\hline
\begin{tabular}[c]{@{}l@{}} \textbf{Title}: fishing \textbf{Storyline}: fish $\rightarrow$ lake $\rightarrow$ kids \\
$\rightarrow$ caught $\rightarrow$ home
\end{tabular} \\ \hline
\begin{tabular}[c]{@{}l@{}}The kids were great at catching fish. \\They woke up early and packed up their tackle\\ box and hiked to the lake. The kids set up their \\lures and caught as many as they could. The \\fish were all caught and the kids laughed \sout{heartily}\\ \textit{[like a group of hyenas]}. They went home and\\ had a great day fishing\end{tabular} \\ \hline
\end{tabular}
\vspace{-1ex}
\caption{\label{table:story} Example of a GPT-2 generated short story on respective title , storyline. We replace the first literal sentence with a generated simile from SCOPE.}
\label{table:example7}
\vspace{-3ex}
\end{table}

\end{document}